# ABHIVYAKTI: A Vision Based Intelligent System for Elder and Sick Persons

Ankit Chaudhary
ankitc.bitspilani@gmail.com

J. L. Raheja
Digital Systems Group
Central Electronics Engineering Research Institute (CEERI)/
Council of Scientific and Industrial Research (CSIR)
Pilani - 333031, Rajasthan, INDIA
jagdish@ceeri.res.in

*Abstract*—This paper describes an intelligent system ABHIVYAKTI, which would be pervasive in nature and based on the Computer Vision. It would be very easy in use and deployment. Elder and sick people who are not able to talk or walk, they are dependent on other human beings and need continuous monitoring, while our system provides flexibility to the sick or elder person to announce his or her need to their caretaker by just showing a particular gesture with the developed system, if the caretaker is not nearby. This system will use fingertip detection techniques for acquiring gesture and Artificial Neural Networks (ANNs) will be used for gesture recognition.

## I. INTRODUCTION

ABHIVYAKTI: it's a word of Hindi Language, used for expressing feelings orally. Old or sick people, who are not able to express their feelings by words or cannot walk, will be the users of this system. During the monitoring of patients, it is possible that nobody is in the room and patients want to eat something or want to call someone, then they can use their hand to make gesture with the developed system and can get what they want. It is assumed that the users are able to move their hand, they are not fully paralyzed. Here in the system, focus of research is on human to machine interaction, in which machine would be able to do action according to the predefined syntax of the gesture made by user. ABHIVYAKTI will have an interface which would include a small smart camera, where users have to show their hand in front of camera. This hand gesture would be interpreted by the system, whether it is valid gesture syntax or not. If the gesture is not included in the rules list, system will not take any action and will give a message of wrong input. If it is valid then the system will work according to predefined action and user will be informed as action done.

ABHIVYAKTI will work on the principles of Computer Vision, where it will use image processing for acquiring gesture and preprocessing. In the proposed system research focus is on 2D systems only because 3D systems are much complex to setup and need at least 2 stereo cameras working synchronously. Also the complexity of modeling of images parameter computation is very high [30], so they are not often used for hand gesture recognition. Also in 3D, it is difficult to measure whether a hand movement was gesture or unintentional [15].

## II. RELATED WORK

Mitra [19] defines gesture recognition a process where user made gesture and receiver recognized them. Many Researchers have done excellent work in this area. Ahn [1] have developed augmented interface table using infrared cameras for pervasive environment. Chaudhary [4] has described designing for intelligent systems in his work. A contour following algorithm has been shown in the figure 1, which extract foreground image from the captured image and further detects hand and face on skin based detection. In gesture recognition color based methods are applicable because of their characteristic color footprint of human skin. The color footprint is usually more distinctive and less sensitive than in the standard (camera capture) RGB color space. Most of color segmentation techniques rely on histogram matching or employ a simple look-up table approach [13][24] based on the training data for the skin and possibly it's surrounding areas. The major drawback of color based localization techniques is the variability of the skin color footprint in different lighting conditions. This frequently results in undetected skin regions or falsely detected non skin textures. The problem can be somewhat alleviated by considering only the regions of a certain size or at certain spatial position.

Another common solution to the problem is the use of restricted backgrounds and clothing like dark gloves or wearing a strip on wrist [16][17][18][20]. Wu [31] has been extracted hand region from the scene using segmentation techniques. Vezhnevets [28] describes many useful methods for skin modeling and detection. Skin color detection and boundary extraction are two important part of gesture extraction form the image. Gesture recognition is the phase in which the data analyzed from the visual images of gestures is recognized as a specific gesture. In this step we assume that gesture image has been extracted from the image captured and now it is target data for gesture recognition. Graph matching is widely used for object mapping in images, but it faces problems in dependency on segmentations [23]. Identification of a hand gesture can be done in many ways depending on the problem to be solved [20].

The interpretations of gestures require that dynamic or static configurations of human hand be measurable by the machine. First attempts to solve this problem resulted in mechanical devices called as glove-based devices that directly measure hand joint angles and spatial position [3][10][29]. In this system requires the user to wear a cumbersome device and carry a load of cables that connect the device to a computer. This hinders the ease and naturalness with which user can interact with computer controlled environment. Potentially, any awkwardness in using gloves and other devices can be overcome by video-based noncontact interaction techniques identifying gestures.

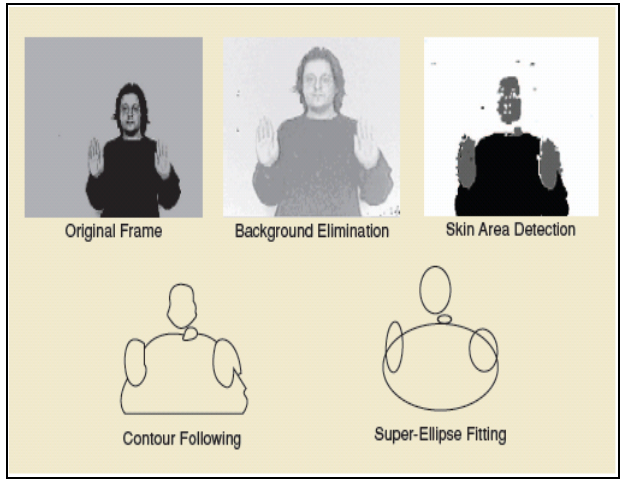

Figure 1. Target image extraction [23]

Most of the static models are meant to accurately describe the visual appearance of the gesturer's hand as they appear to a human observer. To perform recognition of those gestures, some type of parameter clustering technique stemming from vector quantization (VQ) is usually used. Briefly, in vector quantization, an n-dimensional space is partitioned into convex sets using n-dimensional hyper planes, based on training examples and some metric for determining the nearest neighbor. Parameters of the model can be chosen especially to help with the recognition as they did in [8][26]. Structures like cylinders, spheres, ellipsoids and hyper-rectangles are often used to approximate the shape of gesture made like hand or finger joints, hand structure [6]. One approach is to build a classifier such as SVM as in [5][31]. First of all in the captured images which are having hand positions, we have to define the hand structure.

In complex systems such as GREFIT [21][22], the hand model is defined by a set of links and joints. Another technique is learning based methods which can be constructed any intelligent method. Few researchers use fingertip positions as parameters to construct hand image [Table 1]. This approach is based on the assumption that position of fingertips in the human hand relative to the palm, is almost always sufficient to differentiate a finite number of different gesture [2][9][16]. Nguyen [20] uses fingertips recognition and detection based on applying a learning model, to reconstruct the hand model. Fingers are the key points in hand gesture recognition modeling, in the human hand model. The palm must be assumed to be rigid. Kerdvibulvech [12] uses a Gabor feature vector to extract the positions of a guitar player. Other approaches to get the open finger position information include (r, $\theta$) and B-Spline curves or using the curvature information by considering flow between arcs and fingers in the hand model. A different way to detect fingertips is to use pattern matching techniques as templates and can be enhanced by using additional image feature like contours [25].

TABLE 1. Modeling Techniques in different applications

| S.No | Application | Gesture Modeling Technique |
|---|---|---|
| 1 | Fingerpaint [7] | Fingertip template |
| 2 | Finger pointer [11] | Heuristic detection of pointing system |
| 3 | Window manager [14] | Hand pose detection using neural networks |
| 4 | Automatic robot instruction [27] | Fingertip positions in 2D |

## III. PLAN OF WORK

A major motivation for this research on gesture recognition is the potential to use hand gestures in various application, aiming at a natural interaction between the human and various computer controlled displays. Although the current progress in gesture recognition is encouraging, further theoretical as well as computational advancements are needed before gestures can be widely used for human computer interaction. The problem of accurate recognition of postures which use model parameters that cluster in non-convex sets can also be solved by selecting nonlinear clustering schemes. Artificial Neural Networks are one such option, although their use for gesture recognition has not been fully explored [14]. The detection of human fingertips in the human hand structure is an important issue in most hand model studies and in some gesture recognition systems [20]. It is also clear form literature that the work done in this area is not much sufficient and requires keen and new investigation in this direction. This area has many possibilities in the field of Computer Vision and Human machine interaction. Gesture recognition could be probability based if the background is not static or image has other same kind of objects that would also lead to inefficient results.

Generally there are three phases in gesture recognition process - image capture, gesture extraction and gesture recognition. These phases include several issues such as algorithm design, processing speed, system architecture and video interface [23]. Cameras can detect people and recognize their activities in an application environment such as room, a plane, a car or a security checkpoint. The results of these cameras can be analyzed and used to control the operation of devices in these application environments. For the simplicity we will focus only on 2D static gesture recognition. We are making few assumptions here that we will have in our work

(1) We have captured the image using a high resolution camera
(2) Input frame has the size and format that best fits the needs to our requirements.

it and send a signal to transceiver nearby and it will forward the signal further to the rescue team in the control room. In this scenario an advanced version of ABHIVYAKTI, which should be GPS enabled to detect the exact location of the user in the mine. It will work in the case of suffocation, gas poisoning or fire hazards where the person is not able to shout or to tell even a single word.

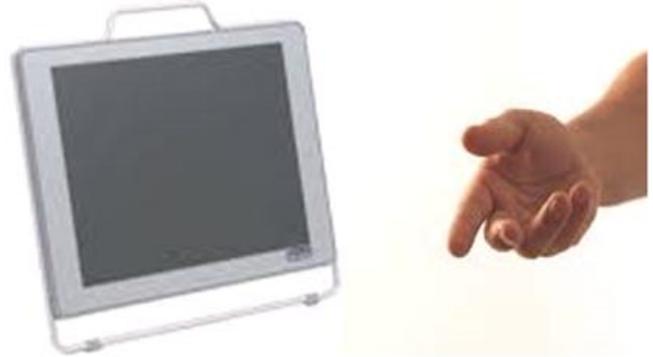

Figure 3. Ease of use

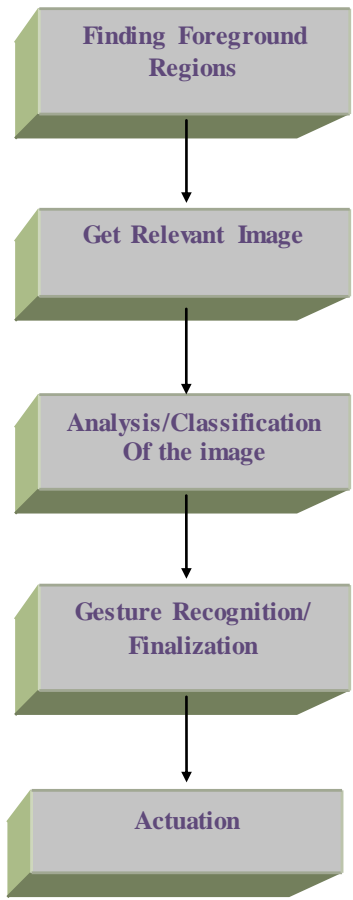

Figure 2. ABHIVYAKTI Methodology

## IV. FUTURE DIRECTIONS

In the future current research can be applied in the field of Disaster management so that lives of more people could be saved. In the damage situations i.e. in mine or building collapse where the human would be under the pile of dust or coal and if he is alive, still he is not able to communicate to other people or to rescue team, for saving his life. This system will have to be on the body of each miner or persons staying or working in such dangerous places and will put the transceiver on the nearby wall. So in any case of disaster, if the user is in danger and cannot go to ground level, he can show predefined gesture syntax to system that will interpret

ACKNOWLEDGMENT

Author would like to thank Director, Central Electronics Engineering Research Institute (CEERI), Pilani, for his support and providing research and experimental facilities.